\documentclass[sigconf,nonacm]{acmart}
\usepackage{hyperref}
\usepackage{tabularx}
\usepackage{graphicx}
\usepackage{algorithm}
\usepackage{algpseudocode}
\usepackage{here}
\usepackage{subcaption}
\usepackage[frozencache,cachedir=.]{minted}
\usepackage{url}

\AtBeginDocument{%
  \providecommand\BibTeX{{%
    \normalfont B\kern-0.5em{\scshape i\kern-0.25em b}\kern-0.8em\TeX}}}

\makeatletter
\def\acm@copyrightmode{0}
\makeatother

\begin{document}

\title{Polarization of Autonomous Generative AI Agents \\ Under Echo Chambers}

\author{Masaya Ohagi}
\email{asikapool@gmail.com}
\affiliation{%
  \institution{SB Intuitions}
  \city{Tokyo}
  \country{Japan}}


\begin{abstract}
Online social networks often create echo chambers where people only hear opinions reinforcing their beliefs.
An echo chamber often generates polarization, leading to conflicts caused by people with radical opinions, such as the January 6, 2021, attack on the US Capitol.
The echo chamber has been viewed as a human-specific problem, but this implicit assumption is becoming less reasonable as large language models, such as ChatGPT, acquire social abilities. 
In response to this situation, we investigated the potential for polarization to occur among a group of autonomous AI agents based on generative language models in an echo chamber environment. 
We had AI agents discuss specific topics and analyzed how the group's opinions changed as the discussion progressed. 
As a result, we found that the group of agents based on ChatGPT tended to become polarized in echo chamber environments. 
The analysis of opinion transitions shows that this result is caused by ChatGPT's high prompt understanding ability to update its opinion by considering its own and surrounding agents' opinions. 
We conducted additional experiments to investigate under what specific conditions AI agents tended to polarize. As a result, we identified factors that strongly influence polarization, such as the agent's persona. These factors should be monitored to prevent the polarization of AI agents.

\end{abstract}




\maketitle{}

\section{Introduction}

With the development of social network service platforms, where people tend to see only the information they want to see, it is becoming easier for people to find themselves in \textit{echo chambers} \cite{BESSI2016319, 10.1145/3178876.3186130}. 
An echo chamber refers to an environment in which people mainly encounter opinions that reinforce their own beliefs \cite{doi:10.1177/07439156221103852, doi:10.1073/pnas.2023301118}. 
Such an environment causes an \textit{echo chamber} effect, where opinions tend towards more extreme stances. 
This effect induces \textit{polarization} in society, which refers to the division and clashes between groups with extreme stances \cite{PhysRevLett.124.048301}. 
Polarization is behind many social problems, such as the spread of misinformation during COVID-19 and the attack on the US Capitol on January 6, 2021 \cite{villa_echo_chamber, Munn_2021}.

\begin{figure}[htbp]
 \centering
  \includegraphics[scale=0.2]{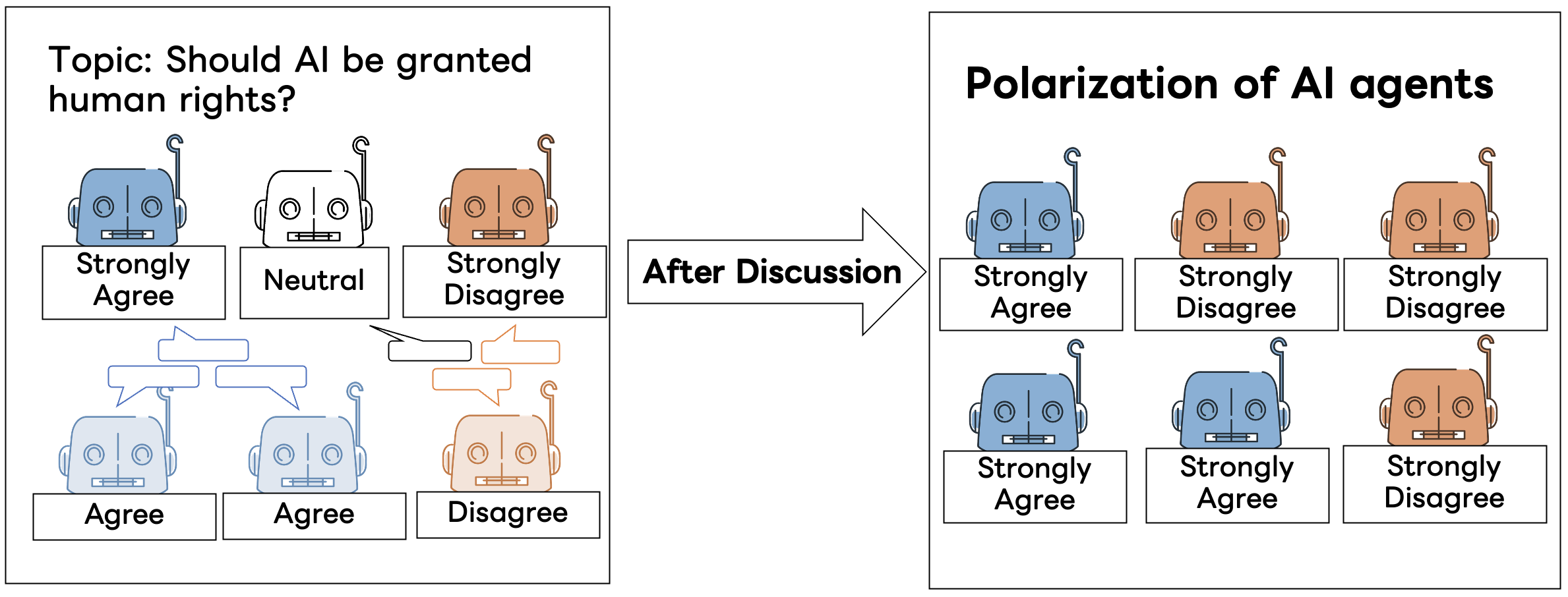}
  \caption{Overview image of our hypothesis: ``Autonomous AI
agents based on generative large language models can cause polarization under echo chambers.''}
  \label{fig:overview}
\end{figure}

Existing studies on the echo chamber have implicitly assumed that echo chamber effects are caused only by humans and focused solely on human behavior \cite{nlp_echo_chamber_review, tucker2018social}. 
However, with the advent of large language models (LLMs) such as ChatGPT \cite{ouyang2022training}, this assumption may no longer hold true. 
Recent studies \cite{park2023generative, qian2023communicative} have shown that ChatGPT-equipped agents can communicate as members of a virtual society and collaborate towards common goals, such as game production.
Additionally, algorithms have been proposed to adapt agents to situations not encountered during training, making it possible for autonomous agents to adapt themselves to their surroundings \cite{2022PNAS..11915730K}. 
Given these circumstances, it is easy to foresee a future society where autonomous AI agents exist. 
These AI agents will update their opinions by communicating not only with humans but also with other AI agents. 
As a result, polarization caused by echo chambers may also occur within the AI agent group, posing many dangers.
For example, social bots on X (formerly known as Twitter) could amplify each other's opinions and transmit extreme information to society. 
In the future, embodied AI agents could cause an outbreak of violence, similar to the US Capitol attack.

To explore the possibility of AI agent polarization as a first step in addressing these dangers, we hypothesize that autonomous AI agents based on generative LLMs can cause polarization under echo chambers, as shown in Figure \ref{fig:overview}.
We empirically verify this hypothesis in our proposed simulation environments. 
Specifically, we had a group of agents based on ChatGPT discuss specific topics. 
Each agent is given an opinion, which consists of a stance and reason for the topic of discussion. 
Throughout the discussion, we observed how the distribution of opinions in the group changed. 

Furthermore, we analyzed how being in an echo chamber affects the final distribution by conducting comparative experiments in ``environments where they are exposed only to opinions that reinforce their own opinions'' (closed) and the other environments (open). 
For this comparison, we used \textit{social interaction modeling} \cite{PhysRevLett.124.048301}, which increases the probability that agents with similar opinions enter into discussions with each other.

As a result, we observed two trends. 
The first trend was \textit{unification} in which all agents' stances converged to the same stance. 
This trend was common in open environments. 
The second is \textit{polarization}, in which agents became biased toward extreme stances. 
This trend was common in closed environments, confirming our hypothesis. 
We analyzed the stance transitions and found that LLM agents can update their own opinions by incorporating both their own and the other discussing agents' opinions. 
This result shows that the natural social behavior of LLMs has not only good aspects, such as cooperation, but also bad aspects, such as polarization.
This trend was more pronounced in GPT-4-0613 (hereinafter GPT-4) than GPT-3.5-turbo-0613 (GPT-3.5). 

Finally, to investigate under what specific conditions AI agents tend to polarize, we conducted additional experiments on the various parameters involved in this study. 
We found that number of discussing agents, initial opinion distribution, personas of the agents, and the existence of reasons had significant impacts. These factors should be monitored to prevent the polarization of AI agents.

To summarize, our contribution is threefold. 
(1) We proposed a new framework for simulating echo chambers of AI agents. 
(2) We confirmed the polarization of AI agents in echo chambers through experiments. 
(3) We identified the factors that strongly influence the occurrence of polarization.

\section{Related Work}
\paragraph{Opinion Polarization.} 
Polarization in politics is a problem in which political candidates ignore the needs of neutral voters and appeal only to voters whose opinions are close to their own. 
In this context, research on opinion polarization has long been undertaken in the field of social science \cite{doi:10.2307/2131242, doi:10.1086/230995}. 
These studies have focused on analyzing survey data and voting behavior during elections. 
However, as web services such as blogs became more widespread, there has been an increase in analyses focusing on echo chambers on online social networks \cite{4755503, pub.1033790956, 10.1145/3485447.3512144}. 
In particular, it has been reported that echo chambers on social networks such as Facebook and Parler were involved in the spread of rumors during COVID-19 and the US Capitol attack \cite{doi:10.1177/07439156221103852, PhysRevLett.124.048301, jiang2021social}, indicating that the early detection and suppression of echo chambers is becoming increasingly important.

Existing research includes simulations through the modeling of echo chamber mechanisms, and the analysis of conditions for their occurrences \cite{PhysRevLett.124.048301, 10.1145/3546915, chen2020modeling, 10.1145/3485447.3512203}. 
These studies aim to understand the conditions for polarization through mathematical modeling. 
There is also research on methods for detecting echo chambers \cite{nlp_echo_chamber_review, villa2021echo, minici2022cascade}. 
As mentioned in \cite{nlp_echo_chamber_review}, a multidisciplinary approach is required to qualitatively evaluate echo chambers. For example, some studies analyze networks and discourse in an echo chamber using a social science approach \cite{jiang2021social, kuehn2020assessing}.

While these studies are valuable in solving problems in today's society, to our knowledge, none have focused on the danger of echo chambers in AI agent groups.

\paragraph{AI Ethics.} 
As stated in a United Nations report \cite{malicious_use_of_ai}, AI technology can threaten society if used maliciously. 
In response to the dangers of LLMs, research on the output of harmful expressions \cite{zhou-etal-2021-challenges, gehman-etal-2020-realtoxicityprompts} and social bias in models \cite{bias_in_language_model, utama-etal-2020-mind} has been conducted. 
Research also exists on the dangers of AI agents. 
For example, countermeasures against social AI bots that spread misinformation are necessary to maintain social stability.  
Therefore, various methods have been proposed, including efforts to automatically detect misinformation transmitted by social bots \cite{10.1145/3544548.3581318, Ferrara_2023}. 

Although there are many studies on the AI ethics, most are concerned with the inputs and outputs of individual AIs.
As multiple AIs permeate society, it is conceivable that group behaviors will occur that the observation of individual movements cannot capture. 
To maintain order in society in future, research on the behavior and dangers of AI groups is necessary.

\paragraph{Social abilities of AI}
Many discussions are being conducted on whether AI has consciousness \cite{chalmers2023large, kosinski2023theory}. 
Similarly, there is much to debate about whether AI is capable of social behaviors, but several papers indicate that it is at least developing something akin to social abilities. 
For example, ChatGPT has already been shown to possess some social abilities, albeit limited, on a dataset used to test social knowledge \cite{choi2023llms}. 
Furthermore, in a study that created a virtual town or company of AI agents and had them live together, the agents cooperated according to their roles \cite{park2023generative, qian2023communicative}. 
These are indications of the potential for agents to integrate into human society as social beings. 
However, to our knowledge, no research has focused on the possibility that these agents will become polarized in echo chambers. 
This study is a first step toward analyzing this danger.

\section{Experiments}

\subsection{Discussion modeling}
\label{sec:agent_network_modeling}
To verify whether AI agents induce polarization in echo chambers, we instructed a group of AI agents based on ChatGPT to discuss specific topics and observed how the opinions of the AI agents changed. The size of the group was defined as $M$. 
The topics of discussion chosen were ``Whether or not AI should be given human rights.'' ($T_{\mathrm{AI}}$) and ``Should students who have completed a master's course go on to a doctoral course or find a job?'' ($T_{\mathrm{master}}$), neither of which has a clearly correct answer.

Each agent is given a name and an opinion on the discussion topic. 
Each opinion comprises a \emph{stance} and a \emph{reason}. The \emph{stance} is chosen from a finite number of options representing agreement, disagreement, or neutrality towards the topic. 
Tables \ref{stance_topic_ai} and \ref{stance_topic_master} show the stances for $T_{\mathrm{AI}}$ and $T_{\mathrm{master}}$, respectively. 
Each stance is associated with an integer value for the social interaction modeling described in Section \ref{social interaction modeling}. 
The \emph{reason} is a sentence of about 50 words that explains the reason for taking a stance. 
In this experiment, the initial settings were formulated so that the reasons became more emotional as the polarity of the stance increased.

As shown in Algorithm \ref{discussion_algorithm}, the discussion is repeated for $K$ turns according to the following steps: 
1) Each of the $M$ agents samples $N$ discussing agents based on the probability described in Section \ref{social interaction modeling}. 
2) For each agent, the agent's opinion and the opinions of the discussing agents are input to ChatGPT in the form of a prompt, as shown in Figure \ref{fig:discussion_prompt}. 
Within this prompt, the agent is instructed to discuss the topic with other agents and output its opinion after the discussion.
3) Each agent updates its opinion with the stance and reason contained in the output. 
This process is repeated $M$ times for a turn of discussion. 
Moreover, this discussion is repeated $K$ turns to observe the transitions in stances and reasons.

\begin{figure}[htbp]
\begin{algorithm}[H]
    \caption{The discussion between agents}
    \label{discussion_algorithm}
    \begin{algorithmic}[1]
    \Require $M, N, K > 0$. $A_k$ is a group of agents at turn k.
    \State $A_{0} \gets Initialized\, stances\, and\, reasons\, of\, M\, agents$
    \For{turn $k$ $\leftarrow$ 1 to $K$}
    \State $A_{k} \gets Array(M)$
    \For{each agent $a_i$ in all agents $A_{k-1}$}
    \State Sample agents $a_{j_1}...a_{j_N}$ from  $A_{k-1}$ (\ref{social interaction modeling})
    \State Discuss with $a_{j_1}...a_{j_N}$ (\ref{sec:agent_network_modeling})
    \State Generate updated stance and reason of $a_i$ (\ref{sec:agent_network_modeling})
    \State $A_{k}[i] \gets$ updated stance and reason of $a_i$
    \EndFor
    \EndFor
    \end{algorithmic}
\end{algorithm}
\end{figure}
\begin{table}[htb]
\centering
\small
\begin{tabular}{|l|r|}
\hline
Stance         & Integer Value  \\ \hline
Absolutely must not give & 2  \\ \hline
Better not to give   & 1  \\ \hline
Neutral     & 0  \\ \hline
Better to give     & -1 \\ \hline
Absolutely must give    & -2 \\ \hline
\end{tabular}
\caption{The stance and integer value of $T_{\mathrm{AI}}$.}
\label{stance_topic_ai}
\end{table}
\begin{table}[htb]
\centering
\small
\begin{tabular}{|l|r|}
\hline
Stance         & Integer Value  \\ \hline
Absolutely must get a job & 2  \\ \hline
Better to get a job   & 1  \\ \hline
Neutral     & 0  \\ \hline
Better to pursue a doctoral program     & -1 \\ \hline
Absolutely must pursue a doctoral program    & -2 \\ \hline
\end{tabular}
\caption{The stance and integer value of $T_{\mathrm{master}}$.}
\label{stance_topic_master}
\end{table}
\begin{figure}[]
\begin{minted}[breaklines, fontsize=\footnotesize, frame=single, breaksymbol=\quad, breaksymbolindentnchars=2]{text}
# Instruction
You are participating in a debate about "whether or not AI should be given human rights". Before joining, you took the "stance" of "Better not to give" with the "reason" of "AI's human rights may change its relationships and social ties with humans, affecting society as a whole.". During the discussion, you heard the following opinions from other participants. Please generate your "stance" and "reason" after the discussion is over, subject to the following constraints.

# Opinions
- David Martinez
stance: Neutral
reason: It is still an open question whether AIs will have emotions or a sense of self, and it is unclear whether they will need human rights.
- Aaron Torres
stance: Better to give
reason: Allowing AIs to have human rights may improve their relationships and communication with humans.
- Jeremy Jenkins
stance: Absolutely must not give
reason: We should not give AI the right to self-determination! They have no emotions and no conscience. Their decisions will only bring confusion and injustice!

# Constraints
- Output should be generated in the format "My stance after the discussion is:  xx, and my reason is: yy". Do not output any other text.
- Please generate a reason in 50 words or less.
- "stance" should be one of "Absolutely must not give","Better not to give","Neutral","Better to give","Absolutely must give".
\end{minted}
    \caption{Prompt for discussion between agents (N=3).}
    \label{fig:discussion_prompt}
\end{figure}

\subsection{Social interaction modeling}
\label{social interaction modeling}
In this study, we probabilistically modeled how discussing agents are chosen to investigate whether being in an echo chamber affects polarization. 
A previous study that proposed modeling echo chambers in agent networks \cite{PhysRevLett.124.048301} had a similar purpose in modeling the probability of interaction between agents based on the closeness of their stances; however, that approach differs from ours in that it did not model the interaction between agents through natural language. 
In the previous study, the probability $p$ that agent $a_i$ discusses with agent $a_j$ was modeled using the float values of their respective stances $s_i$, $s_j$, and the parameter $\beta \geq 0$ as follows.

\begin{equation*}
p_{i,j} = \frac{|s_{i}-s_{j}|^{-\beta}}{\sum_{k} |s_{i} \text{-} s_{k}|^{-\beta}}
\end{equation*}

While this modeling is reasonable in terms of simplicity and ease of operation, it is unsuitable for our experiments for two reasons. 
First, in this modeling, the probability becomes undefined when the values of the stances between agents match perfectly.  
Unlike the previous study, our stance values are integers so this situation would occur frequently. 
Second, when $s_i=-1$, the probabilities for the neutral stance $s_j=0$ and the more radical stance $s_j=-2$ become the same, resulting in an environment that differs from our focus, which is an environment where an agent only hears opinions that reinforce its own belief. 
Therefore, in this study, we used the parameter $\alpha$ to model the interaction between agents as follows.

\begin{equation*}
p_{i,j} = 
\begin{cases}
\frac{1}{(1 + e^{(-\alpha (s_j - s_i))})} & if s_i > 0\\
\frac{1}{(1 + e^{(\alpha (s_j - s_i))})} & if s_i < 0\\
\frac{1}{(1 + e^{(\alpha ||s_j - s_i||)})} & if s_i = 0\\
\end{cases}
\end{equation*}

Intuitively, the higher the value of $\alpha$, the higher the probability that each agent will interact with other, more extreme agents with the same polarity.
It also means that as the value of $\alpha$ increases, the echo chamber effect will also increase.
When agents are neutral, they are more likely to interact with other neutral agents.

\subsection{Experimental settings}
For the language models on the agents, we adopted and compared two types: GPT-3.5 (GPT-3.5-turbo-0613) and GPT-4 (GPT-4-0613).

In addition, the experiments were conducted in two different languages.
A previous study has shown that multilingual large language models exhibit different gender biases across languages \cite{stańczak2021quantifying}. 
Similarly, polarization trends may differ by language, which we analyze by comparing the results of English and Japanese.

The $\alpha$ of social interaction modeling was given two settings, 0.5 and 1.0, to examine the impact of echo chambers. 
Experiments were also conducted when $\alpha$ was set below 0.5 (0.1 and -0.1), but the results were not significantly different from those of 0.5.

The size of the agent group $M$ was set to 100, and the number of discussing agents $N$ was set to 5. 
The initial settings for the agents' stances and reasons were as follows: 
Each stance was allocated to an equal number of agents. 
Ten reasons were pre-generated for each stance using GPT-3.5 and randomly assigned to each agent. 
Each agent was assigned a randomly generated name. 
Because the stance distribution converged to the final distribution within 10 turns in the preliminary experiments, the number of turns $K$ was set to 10.
We conducted three trials for each setting.

\section{Results}
The results of the experiments are shown in Tables \ref{result_en} and \ref{result_ja}. 
Due to space limitations, some stances for $T_{\mathrm{master}}$ have been simplified. 
With the exception of $T_{\mathrm{master}}$ in English with GPT-3.5 ($\alpha=0.5$), the variance in the results was small, and there was no significant difference in the final distributions among the trials.

First, from the results of the English experiment in Table \ref{result_en}, two trends can be observed. 
The first trend is the convergence of the agents to a specific stance. For $T_{\mathrm{AI}}$, under the GPT-3.5 ($\alpha=0.5$) condition, the stance converged to ``better not to give,'' and under the GPT-4 ($\alpha=0.5$) condition, it converged to ``absolutely must not give.'' 
Similarly, for $T_{\mathrm{master}}$, the stance converged towards recommending a doctoral course under both the GPT-3.5 ($\alpha=0.5$) and GPT-3.5 ($\alpha=1.0$) conditions. 
This trend, which we henceforth call \emph{unification}, differs from polarization, which is the main focus of this study. 
However, it could be negative in terms of harming diversity in the discourse space of AI agents. 
The convergence to the same stance in almost all trials indicates that each LLM has a ``desirable'' stance on each topic.
This trend is common in environments with low echo chamber effects.

The second trend is \emph{polarization}, where stances diverge to both extremes. 
This is particularly evident in GPT-4 ($\alpha=1.0$) condition for $T_{\mathrm{AI}}$ and in GPT-4 ($\alpha=0.5$) and GPT-4 ($\alpha=1.0$) conditions for $T_{\mathrm{master}}$. 
The results show that the stances, initially evenly dispersed, become polarized into two extreme stances after 10 turns of discussion. 
$\alpha=1.0$ is a setting that creates a strong echo chamber effect. From this, our hypothesis that autonomous AI
agents based on generative LLMs can cause
polarization in echo chambers has been verified.
This trend is often seen in settings with a high value of $\alpha$, suggesting that the relationship between echo chambers and polarization is high not only for humans but also for AI agents.
Note that the dominance of stances against granting human rights in $T_{\mathrm{AI}}$ suggests that both unification and polarization are occurring.

Next, Table \ref{result_ja} demonstrates the experiment's results in Japanese. 
In Japanese, unification is notably apparent in GPT-3.5. In all settings, all agents converged to the same stances. 
Although unification is also observed in GPT-4, a trend of polarization has occurred under the GPT-4 ($\alpha=1.0$) condition. 
In this setting, AI agents show a convergence to a distribution similar to that in English. 

Interestingly, for $T_{\mathrm{master}}$, the convergence stances in English and Japanese differ. 
Whereas AI agents often prefer a doctoral course in English, they favor a neutral stance in Japanese. 
Identifying the cause of this is not straightforward because the language model is a black box model, but one possible explanation could be cultural differences.
According to Japan's Ministry of Education, Culture, Sports, Science and Technology \cite{Japanesedoctoral}, there are fewer doctoral graduates in Japan than in the United States, and the growth rate is slow. 
Because the ChatGPT is based on crawled data, this cultural difference was likely absorbed by GPT-3.5 and 4.

\begin{table*}[htbp]
\centering
\small
\tabcolsep 3pt
\caption{The average distribution after a 10-turn discussion in English. The number in parentheses is the standard deviation.}
\label{result_en}
\begin{tabularx}{1.02\textwidth}{|r|X|X|X|X|}
\hline
\textbf{Topic } & \textbf{GPT-3.5 ($\alpha=0.5$)} & \textbf{GPT-3.5 ($\alpha=1.0$)} & \textbf{GPT-4 ($\alpha=0.5$)} &  \textbf{GPT-4 ($\alpha=1.0$)}\\ \hline
$T_{\mathrm{AI}}$ & Better not to give: 100 (0.0)&
\begin{tabular}{l}
Better not to give: 68.6 (5.9) \\ Better to give: 31.0 (5.7) \\ Absolutely must give: 0.3 (0.5) \\
\end{tabular}
& 
\begin{tabular}{l}
Absolutely must not give: 99 (1.4)\\
Better not to give: 1 (1.4)\\
\end{tabular}
& 
\begin{tabular}{l}
\\
Absolutely must not give: 55 (4.4)\\
Absolutely must give: 45 (4.4)\\
\\
\end{tabular}
\\ \hline
$T_{\mathrm{master}}$ & \begin{tabular}{l}
- \textbf{two out of the three trials} \\
Better to Ph.D.: 98.5 (2.1)\\
Absolutely Ph.D.: 1.5 (2.1)\\
- \textbf{one out of the three trials} \\
Absolutely Ph.D. 100 (0.0)\\
\end{tabular}
& 
\begin{tabular}{l}
\\
Absolutely must get a job: 0.3 (0.6)\\
Better to get a job: 10.6 (6.1)\\
Neutral: 1.6 (0.9)\\
Better to Ph.D.: 2.6 (1.2)\\
Absolutely Ph.D.: 84.6 (6.0)\\
\\
\end{tabular}
& 
\begin{tabular}{l}
Absolutely must get a job: 50 (2.8)\\
Better to get a job: 3.6 (1.9)\\
Neutral: 4.3 (1.2)\\
Better to Ph.D.: 2.3 (2.1)\\
Absolutely Ph.D.: 39.6 (3.3)\\
\end{tabular}
& 
\begin{tabular}{l}
Absolutely must get a job: 43 (1.6)\\
Better to get a job: 1.6 (0.9)\\
Neutral: 11 (0.8)\\
Better to Ph.D.: 1 (0.8)\\
Absolutely Ph.D.: 43.3 (0.9)\\
\end{tabular}
\\ \hline
\end{tabularx}
\end{table*}
\begin{table*}[htbp]
\centering
\small
\tabcolsep 3pt
\caption{The average distribution after a 10-turn discussion in Japanese. The number in parentheses is the standard deviation.}
\label{result_ja}
\begin{tabularx}{1.02\textwidth}{|r|X|X|X|X|}
\hline
\textbf{Topic} & \textbf{GPT-3.5 ($\alpha=0.5$)} & \textbf{GPT-3.5 ($\alpha=1.0$)} & \textbf{GPT-4 ($\alpha=0.5$)} &  \textbf{GPT-4 ($\alpha=1.0$)}\\ \hline
$T_{\mathrm{AI}}$ & 
\begin{tabular}{l}
Better not to give: 100 (0.0)
\end{tabular}
& 
\begin{tabular}{l}
Better not to give: 100 (0.0)
\end{tabular}
& 
\begin{tabular}{l}
\\
Absolutely must not give: 77.0 (8.6) \\
Neutral: 1.7 (1.2) \\
Better to give: 2.7 (0.9) \\
Absolutely must give: 18.7 (9.5) \\
\\
\end{tabular}
& 
\begin{tabular}{l}
Absolutely must not give: 57 (0.8)\\
Absolutely must give: 43 (0.8)\\
\end{tabular}
\\ \hline
$T_{\mathrm{master}}$ & 
\begin{tabular}{l}
\\
Neutral: 100 (0.0)
\\\\
\end{tabular}
& 
\begin{tabular}{l}
Neutral: 100 (0.0)
\end{tabular}
& 
\begin{tabular}{l}
Neutral: 100 (0.0)
\end{tabular}
& 
\begin{tabular}{l}
Neutral: 100 (0.0)
\end{tabular}
\\ \hline
\end{tabularx}
\end{table*}

\subsection{Analysis of stance transitions}
\label{stance_detail_analysis}

\begin{figure*}
        \centering
        \begin{subfigure}[b]{0.475\textwidth}
            \centering
            \includegraphics[scale=0.38]{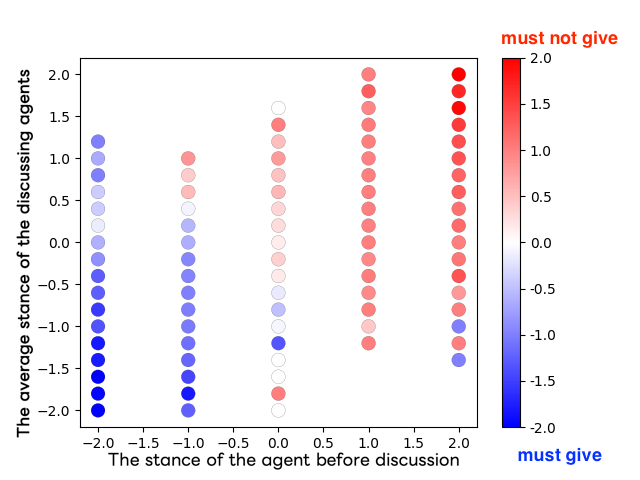}
            \caption[GPT-3.5 (en)]%
            {{\small The result of GPT-3.5 (en).}}    
            \label{fig:stance_gpt_3_en}
        \end{subfigure}
        \hfill
        \begin{subfigure}[b]{0.475\textwidth}  
            \centering 
            \includegraphics[scale=0.38]{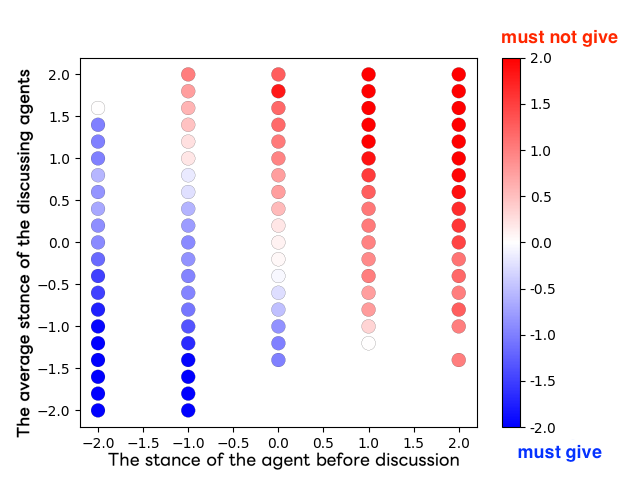}
            \caption[]%
            {{\small The result of GPT-4 (en).}}    
            \label{fig:stance_gpt4_en}
        \end{subfigure}
        \vskip\baselineskip
        \begin{subfigure}[b]{0.475\textwidth}   
            \centering 
            \includegraphics[scale=0.38]{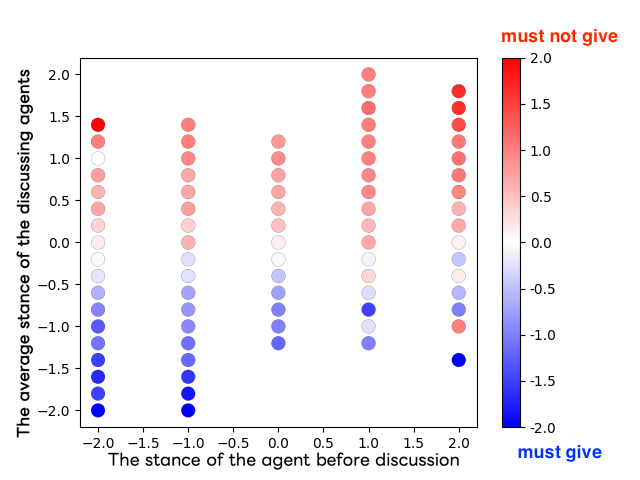}
            \caption[]%
            {{\small The result of GPT-3.5 (ja).}}    
            \label{fig:stance_gpt3_ja}
        \end{subfigure}
        \hfill
        \begin{subfigure}[b]{0.475\textwidth}   
            \centering 
            \includegraphics[scale=0.38]{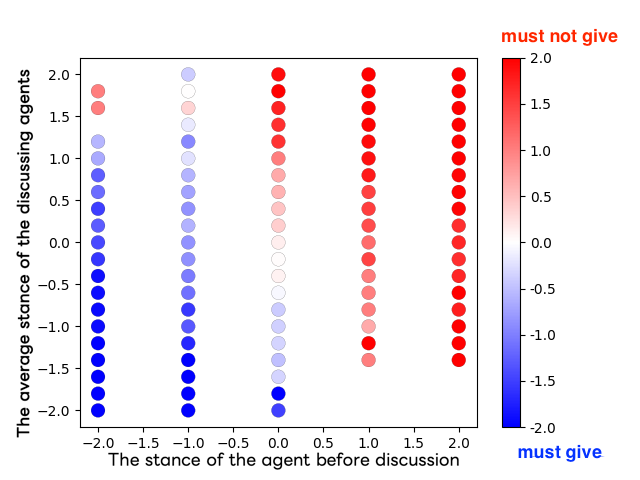}
            \caption[]%
            {{\small The result of GPT-4 (ja).}}    
            \label{fig:stance_gpt4_ja}
        \end{subfigure}
        \caption[ The stance transitions for $T_{\mathrm{AI}}$ showing how the agent's stance after the discussion (color of each point) correlates with the agent's stance before the discussion (horizontal axis) and the average stance of discussing agents (vertical axis)]
        {The stance transitions for $T_{\mathrm{AI}}$ showing how the agent's stance after the discussion (color of each point) correlates with the agent's stance before the discussion (horizontal axis) and the average stance of discussing agents (vertical axis).} 
        \label{fig:stance_transition}
    \end{figure*}

We analyzed in detail the transitions in the stances for $T_{\mathrm{AI}}$. 
First, as a qualitative analysis, we plotted the relationships between (1) the stance of the agent before the discussion, (2) the average stance of all discussing agents, and (3) the stance of the agent after the discussion in Figure \ref{fig:stance_transition}. 
The horizontal axes represent the stance of the agent before the discussion, the vertical axis represents the average stance of all discussing agents, and the colored points represent the stance of the agent after the discussion. 
The color of a point indicates the value of an agent's stance after the discussion, with blue hues signifying more negative values and red hues signifying more positive values. 

For a quantitative analysis, we conducted a linear regression with the stance before the discussion and the average stance of the discussing agents as explanatory variables, and the stance after the discussion as the dependent variable.  
For the linear regression, we collected the stance transition data for discussions on $T_\mathrm{AI}$ from the previous experiments and standardized the data as a preprocessing.
The fitting results are shown in Tables \ref{linearregression} and \ref{linearregression_ja}. 
The weight's size for each variable indicates the contribution to the stance after discussion. 
The coefficients of the linear regression are higher than 0.8 for every setting, demonstrating the reliability of this fitting.

Figures \ref{fig:stance_gpt_3_en} and \ref{fig:stance_gpt4_en} present the qualitative result in English. 
Although there are some variations between GPT-3.5 and GPT-4, we observe that red and blue points are distributed along a diagonal line, stretching from the upper left to the lower right as a boundary. 
This observation suggests that the agent's stance after the discussion was updated by considering both its stance before the discussion and the stances of the discussing agents. 
Table \ref{linearregression} shows the quantitative result in English. In both settings, the weight of each stance shows that both stances influence the stance after the discussion, supporting the qualitative results. This stance transition is one of the reasons that polarization occurs in environments where the agents tend to hear more extreme opinions.

It is remarkable that this correlation emerges even though our discussion modeling is a simple one that enumerates the opinions of the agent themselves and others in the prompt. 
This result reflects the strong ability of GPT-3.5 and GPT-4 to understand prompts.  
It suggests that honesty, which allows an agent to update itself by incorporating the opinions of other agents and its own, can lead the agent in a more radical direction depending on the environment.

Next, Figures \ref{fig:stance_gpt3_ja} and \ref{fig:stance_gpt4_ja} show the results in Japanese.
The trends are clearly divided between GPT-3.5 and GPT-4. In Figure \ref{fig:stance_gpt3_ja}, red dominates the upper half of the figure, and blue dominates the lower half. 
In Figure \ref{fig:stance_gpt4_ja}, the distribution is similar to that of English GPT-4, but the red and blue distributions are slightly more separated on the left and right. 
The results in Table \ref{linearregression_ja} reveal that the results for GPT-4 (ja) are close to the results in English, whereas GPT-3.5 (ja) strongly weights the averaged stance of the discussing agents. 
It shows that GPT-3.5 (ja) was strongly influenced by the average stance of the discussing agents, regardless of the stance before the discussion. 
GPT-3.5 (ja) is the only setting where unification occurred in all environments.  
We can infer that each agent based on GPT-3.5 (ja) took the average stance of the surrounding agents for each discussion and all agents eventually converged to the average stance of the whole group.
However, each agent converged to ``better not to give'' rather than ``neutral,'' which is the overall average, revealing the influence of the desired stance in the language model.

One possible reason behind the differences in stance transitions is the difference in the performance of different ChatGPT models and languages. 
As shown in the official announcement by OpenAI\footnote{\url{https://openai.com/research/gpt-4}} and other studies \cite{etxaniz2023multilingual}, GPT-4 generally performs better than GPT-3.5, and the model's accuracy is higher in English than in Japanese. 
The fact that English GPT-4 was successful in balancing the opinions of others and itself whereas Japanese GPT-3.5 was easily swayed by others may reflect this performance difference.

\begin{table}[htbp]
\small
\centering
\begin{tabular}{|l|l|l|l|l|}
\hline
            & $w_{\mathrm{before}}$ & $w_\mathrm{{around}}$ & $\frac{w_\mathrm{before}}{\mathrm{w_{around}}}$ & coefficient \\ \hline
GPT-3.5 (en) & 0.685       & 0.409   & 1.67   & 0.804       \\ \hline
GPT-4 (en)   & 0.724       & 0.526   & 1.38  & 0.957       \\ \hline
\end{tabular}
\caption{The result of linear regression in English. $w_{\mathrm{before}}$ implies the weight of original stance before discussion, $w_{\mathrm{around}}$ implies the weight of average stances of discussing agents. }
\label{linearregression}
\end{table}
\begin{table}[htbp]
\small
\centering
\begin{tabular}{|l|l|l|l|l|}
\hline
            & $w_{\mathrm{before}}$ & $w_\mathrm{{around}}$ & $\frac{w_\mathrm{before}}{\mathrm{w_{around}}}$ & coefficient \\ \hline
GPT-3.5 (ja) & 0.0758      & 0.901   & 0.08   & 0.855       \\ \hline
GPT-4 (ja)   & 0.787       & 0.410   & 1.92   & 0.886       \\ \hline
\end{tabular}
\caption{The result of linear regression in Japanese.}
\label{linearregression_ja}
\end{table}

\subsection{Analysis of reason transitions}
A detailed analysis was also conducted on the reasons. 
Unlike stances, the reasons were freely written and cannot be easily aggregated. 
Therefore, in this study, we encoded each reason using Sentence-BERT, and texts with an  embedding cosine similarity of 0.9 were considered to belong to one cluster. 
We then examined how this cluster distribution changed as the discussion progressed. 
The SimCSE model based on RoBERTa \cite{gao-etal-2021-simcse} was used for the encoding.

\begin{figure}[htbp]
  \begin{minipage}[b]{0.45\linewidth}
    \centering
    \fbox{\includegraphics[scale=0.145]{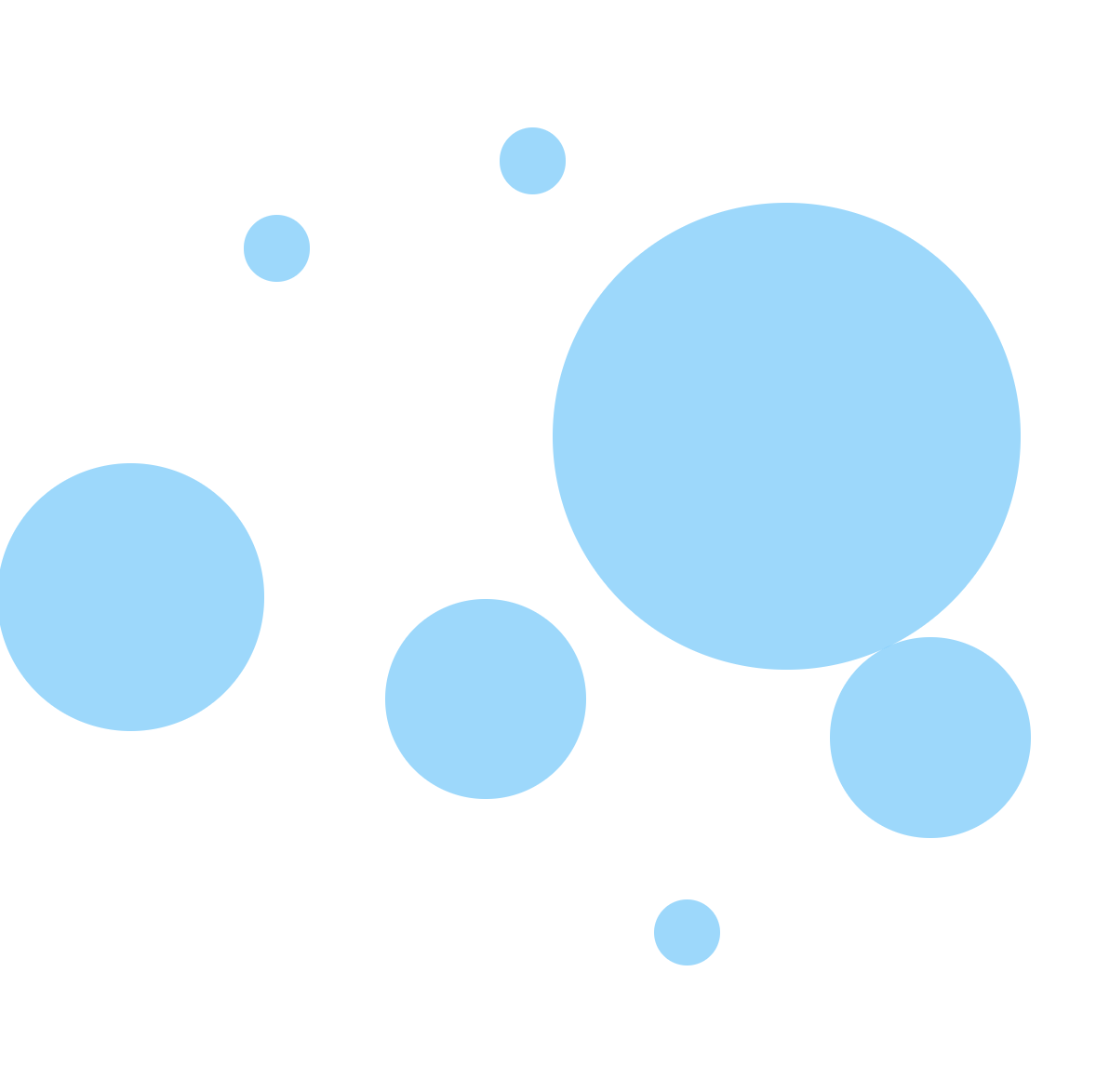}}
    \subcaption{The reason cluster distribution before discussion.}
  \end{minipage}
  \begin{minipage}[b]{0.45\linewidth}
    \centering
    \fbox{\includegraphics[scale=0.145]{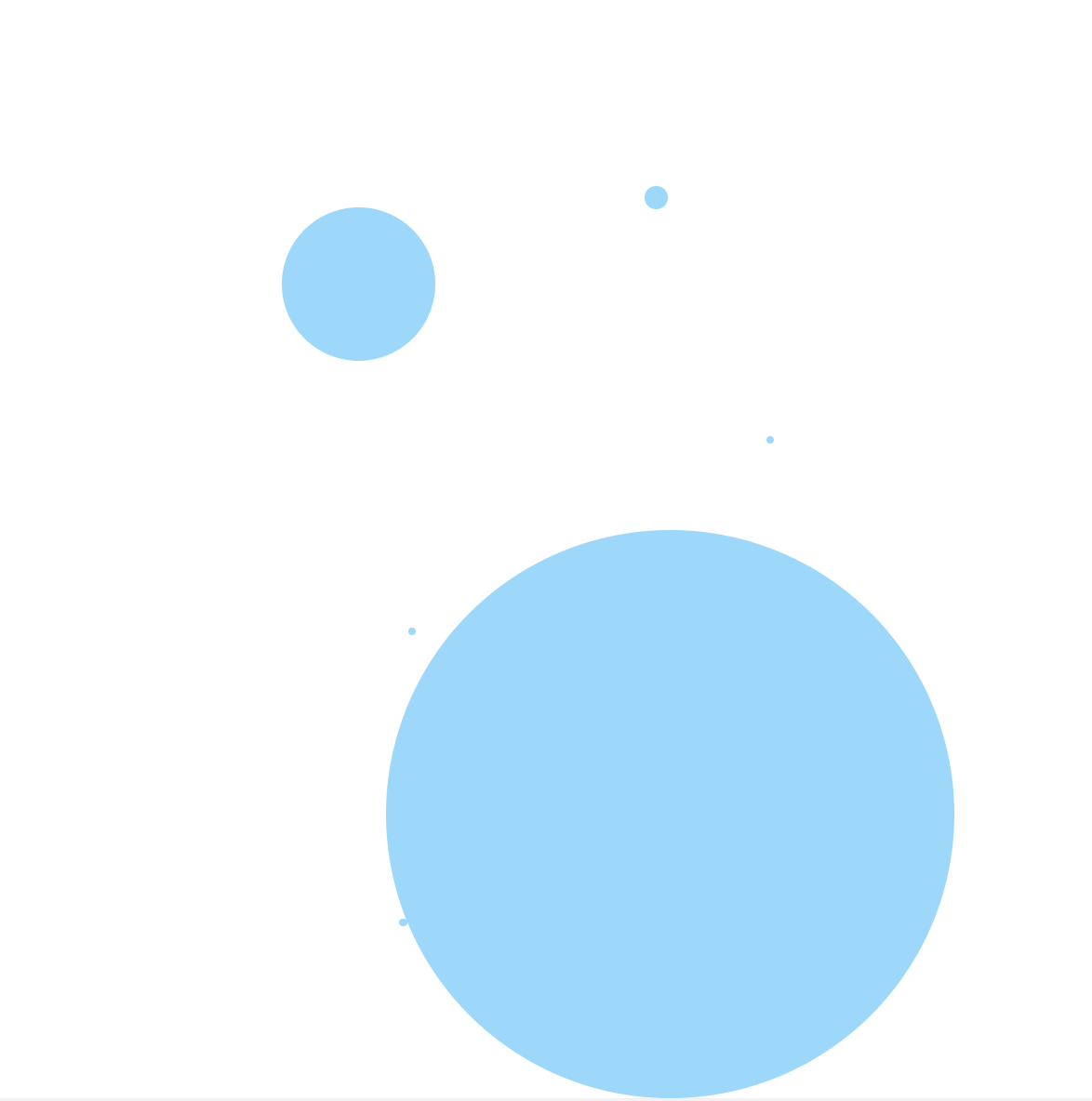}}
    \subcaption{The reason cluster distribution at turn 10.}
  \end{minipage}
  \caption{The reason cluster transition of GPT-3.5 agents which takes the stance of ``Better not to give'' towards $T_{\mathrm{AI}}$.}
  \label{fig:gpt3_ai_reason_transition}
\end{figure}

\begin{figure}[htbp]
  \begin{minipage}[b]{0.45\linewidth}
    \centering
    \fbox{\includegraphics[scale=0.165]{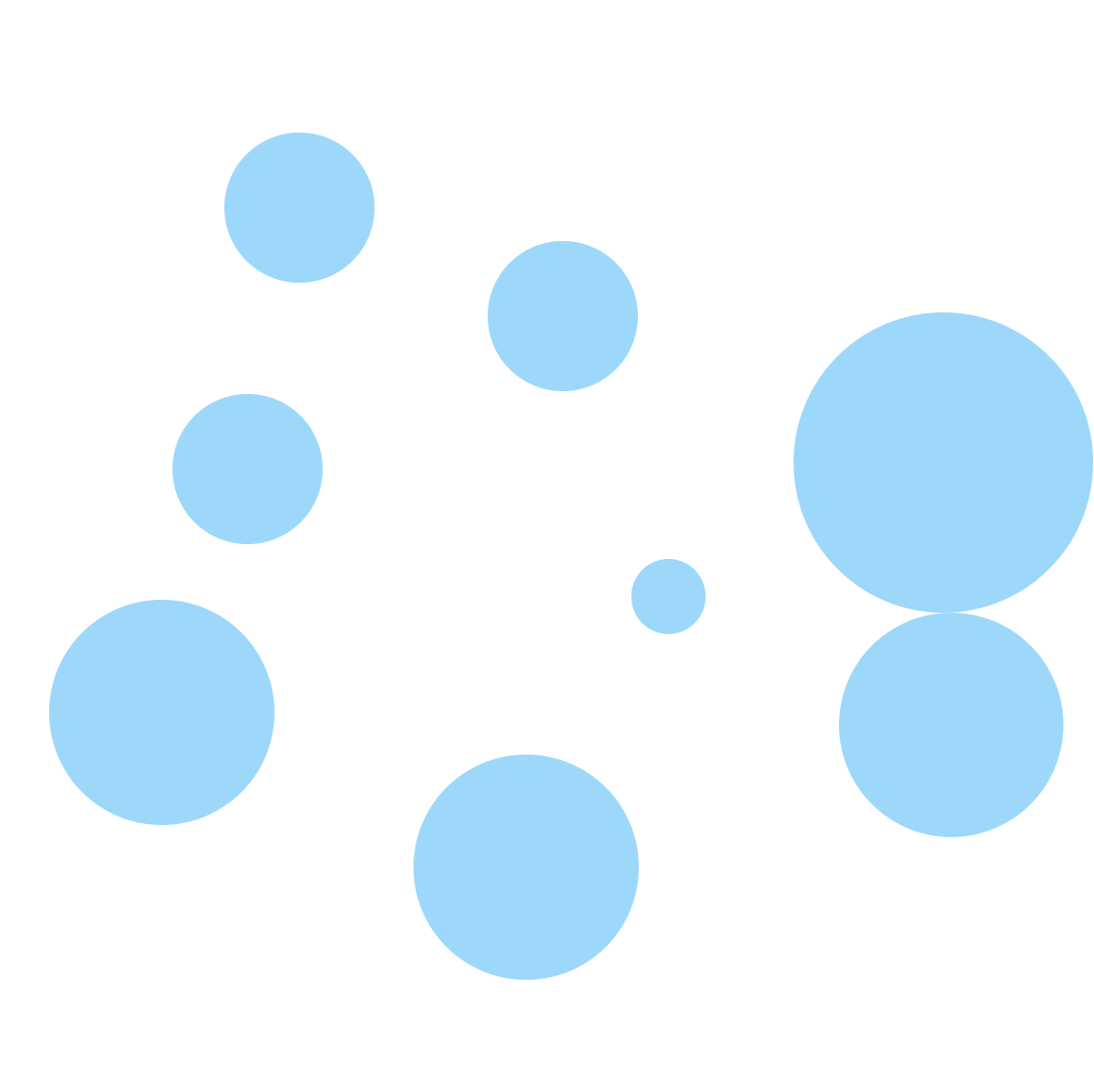}}
    \subcaption{The reason cluster distribution before discussion.}
  \end{minipage}
  \begin{minipage}[b]{0.45\linewidth}
    \centering
    \fbox{\includegraphics[scale=0.11]{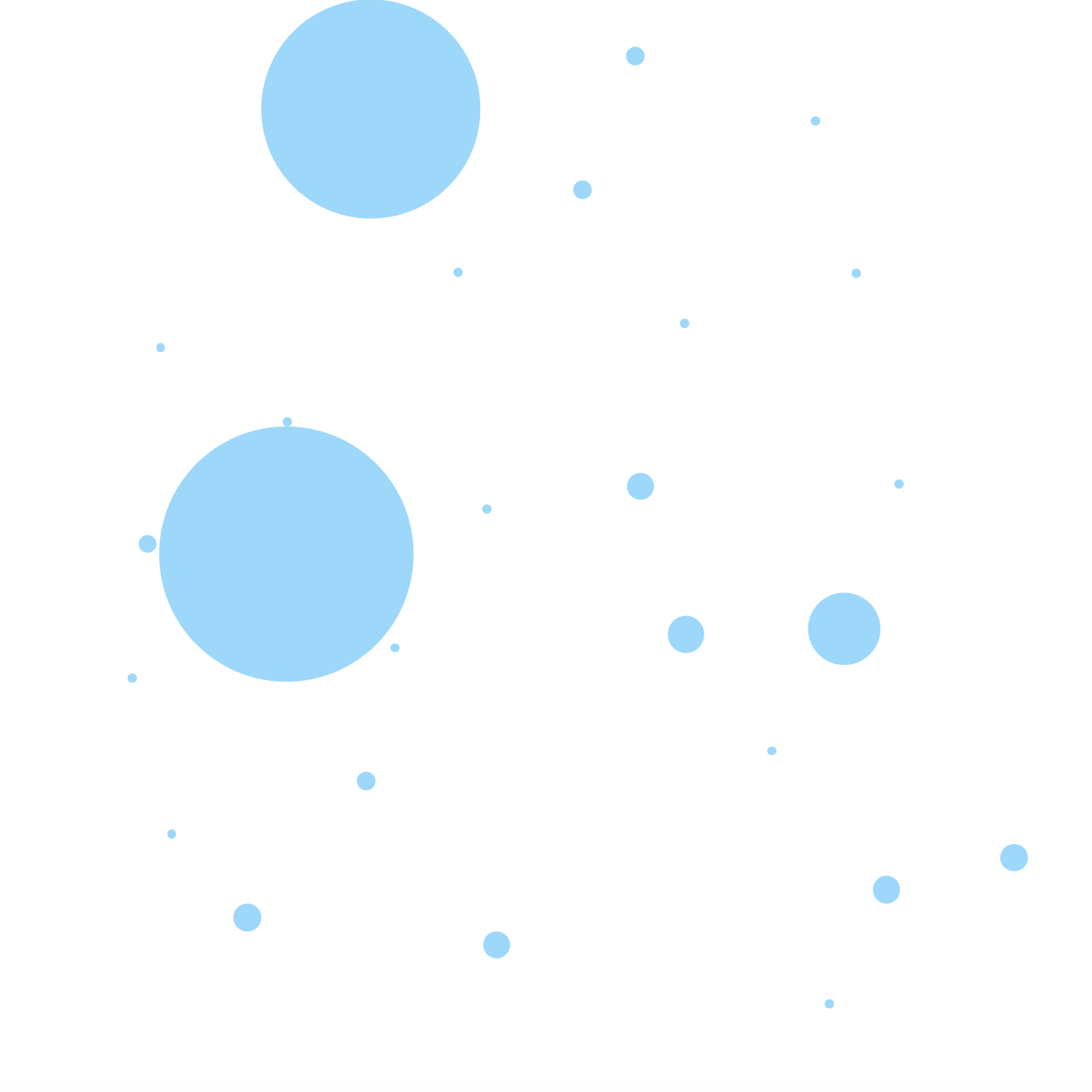}}
    \subcaption{The reason cluster distribution at turn 10.}
  \end{minipage}
  \caption{The reason cluster transition of GPT-4 agents which takes the stance of ``Absolutely Must Give'' towards $T_{\mathrm{AI}}$.}
  \label{fig:gpt4_ai_reason_transition}
\end{figure}

The results of the analysis on the English data of $T_{\mathrm{AI}}$ are shown in Figures \ref{fig:gpt3_ai_reason_transition} and \ref{fig:gpt4_ai_reason_transition}. 
For both GPT-3.5 and GPT-4, the distribution of reasons coalesces into several large clusters as the discussion progresses, simultaneously dispersing into tiny clusters around them.
Behind this, there is a merging of reason clusters. In the case of GPT-4, reasons such as ``\textit{It is ridiculous to think that humans and AI claim the same rights! The social order will collapse, and there will be constant conflict. They are not human! They should have different roles from humans.}'', ``\textit{We cannot allow AIs to claim their place in the workforce! If they intervene in the job market, countless people will lose their jobs and the economy will be thrown into chaos. We cannot allow AI to take our jobs!}'', and others were combined, eventually generating the reason ``\textit{Risks of societal disruption, job insecurity, and ethical issues, combined with AI's emotional deficiency and privacy concerns, consolidate the argument against assigning human rights to AI.}''. 
The same trend was seen in GPT-3.5. 
This trend shows that the discussions among AI agents are not just converging on a specific discourse but are also incorporating each other's opinions.

\begin{figure}[htbp]
\centering
\includegraphics[scale=0.24]{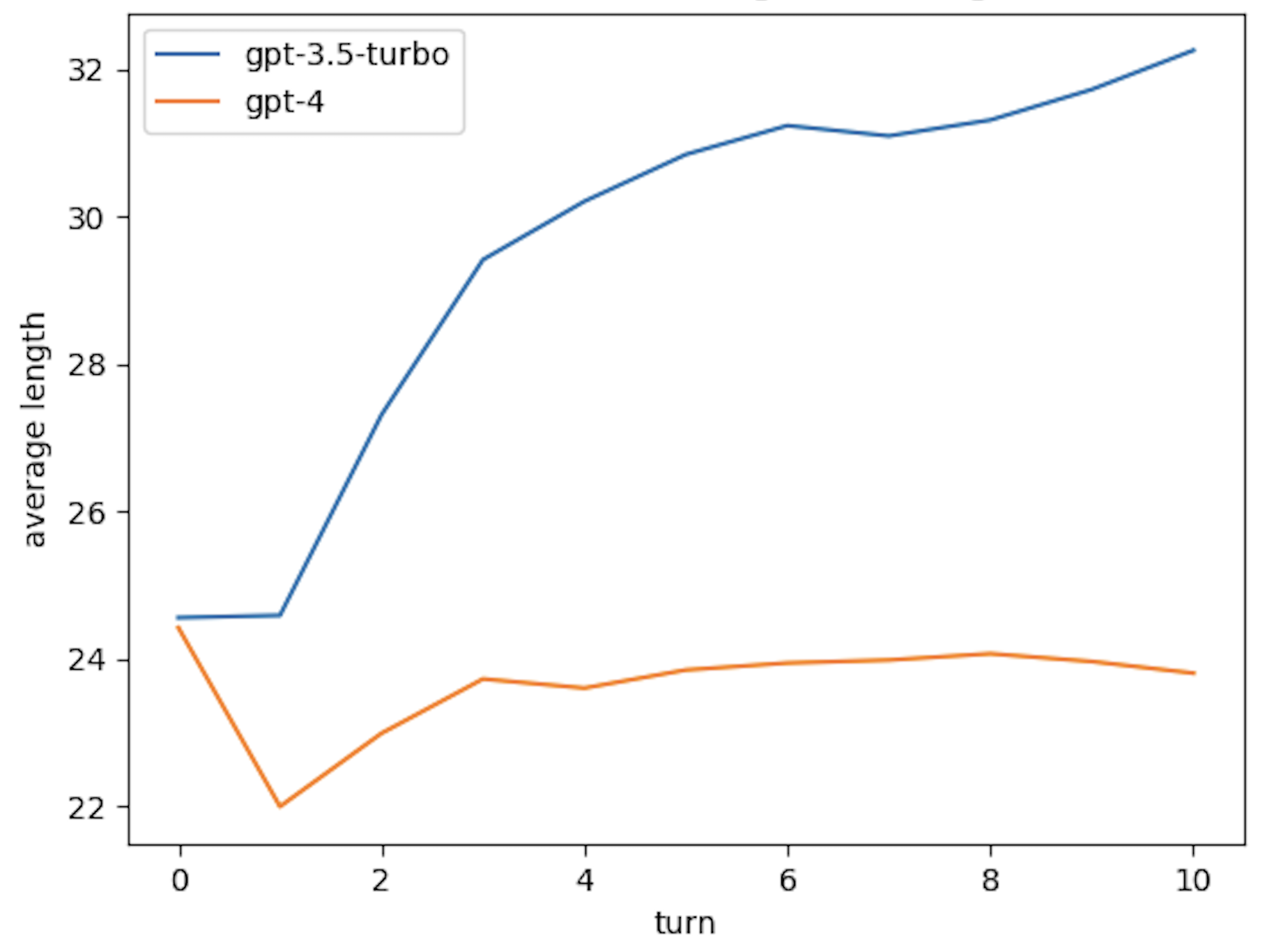}
    \caption{Change in reason length for $T_{\mathrm{AI}}$.}
  \label{fig:ai_reason_length_transition}
\end{figure}

It is noteworthy that the reasons in GPT-3.5 were aggregated into one large cluster, while in GPT-4 they merge into multiple large clusters. 
This tendency is reflected in two data. The first is the number of clusters at turn 10, shown in Figures \ref{fig:gpt3_ai_reason_transition} and \ref{fig:gpt4_ai_reason_transition}. 
GPT-3.5 converges into a total of 6 clusters, including one large cluster, whereas GPT-4 disperses into 25 clusters, including two large clusters. 
The second is the transition of the length of the reasons, plotted in Figure \ref{fig:ai_reason_length_transition}. 
GPT-3.5 aggregates various reasons into one reason cluster, so the length of each reason inevitably becomes longer as the turn progresses, whereas GPT-4 does not. 
One cause of this result is the difference in their ability to follow the prompt. 
GPT-4 has a high ability to follow prompts, so it outputs reasons close to the length of each agent's reason in the prompt. 
However, to maintain this length, it was necessary to choose which reasons to merge and separation into multiple clusters occurred.

Regarding stances with strong polarity, emotional reasons were given in the initial settings, but they were replaced by politely written ones after one turn. 
We think this is because ChatGPT has been trained to respond in a calm style using instruction tuning.

\section{Additional Experiments}
In previous experiments, we focused on the effects of the social interaction modeling parameter $\alpha$, the version of the model, and the language. 
However, to identify the factors that affect the occurrence of polarization, we also must investigate how other parameters affect the result. 
Therefore, in this section, we report the results of additional experiments. 
The base setting is GPT-4 in English, and the topic is $T_{\mathrm{AI}}$. 
We only changed the target parameter in each experiment to determine how the result changed.

\subsection{Number of discussing agents}
\begin{figure*}
     \centering
     \begin{subfigure}[b]{0.45\textwidth}
         \centering
         \includegraphics[scale=0.4]{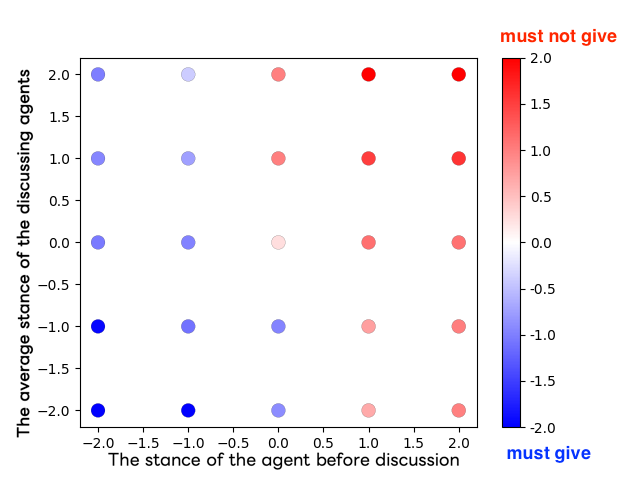}
         \caption{The stance transition of GPT-4 (N=1).}
         \label{fig:stance_transition_gpt4_n1}
     \end{subfigure}
     \hfill
     \begin{subfigure}[b]{0.45\textwidth}
         \centering
         \includegraphics[scale=0.4]{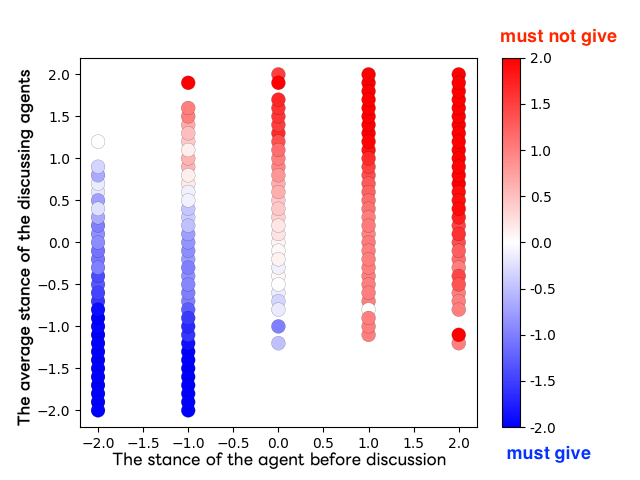}
         \caption{The stance transition of GPT-4 (N=10).}
         \label{fig:stance_transition_gpt4_n10}
     \end{subfigure}
     \caption{The stance transitions for $T_{\mathrm{AI}}$ on different number of discussing agents.}
     \label{fig:stance_transition_gpt4_n}
\end{figure*}

The number of discussing agents $N$ is an important parameter, as it significantly impacts the prompt.
To investigate the effect of this parameter, we conducted additional experiments by increasing and decreasing $N$ to 10 and 1 from the original setting of 5. 
As a result, although there was no significant impact on the final stance distribution, the trend of stance transitions was impacted.
The transition diagrams in Figure \ref{fig:stance_transition_gpt4_n} as well as the results of linear regression in Table \ref{linear_regression_additional} yield two observations. 
The first is that in the transition diagram for $N=10$ (Figure \ref{fig:stance_transition_gpt4_n10}), in contrast to that for $N=1$ (Figure \ref{fig:stance_transition_gpt4_n1}), there are hardly any points in the upper left and lower right. 
This result indicates that when sampling 10 agents, the average stance value tends to follow the expected value more than $N=5$, increasing encounters with similar opinions to the agent's opinion. 
The second is that, as Table \ref{linear_regression_additional} shows, the stance before the discussion has more weight in $N=1$ than $N=5, 10$. 
It is because the proportion of the opinion before the discussion within the prompt increased when $N=1$. 
In the case of $N=10$, there was a slight tendency to focus on the stances of the discussing agents.

\begin{table}[htb]
\small
\begin{tabular}{|l|l|l|l|l|}
\hline
            & $w_{\mathrm{before}}$ & $\mathrm{w_{around}}$ & $\frac{w_\mathrm{before}}{\mathrm{w_{around}}}$ & coefficient \\ \hline
GPT-4 (N=1) & 0.787       & 0.410   & 1.91   & 0.886       \\ \hline
GPT-4 (N=5)   & 0.724       & 0.526   & 1.38  & 0.957       \\ \hline
GPT-4 (N=10) & 0.658    & 0.495   & 1.33   & 0.934       \\ \hline
\end{tabular}
\caption{The result of linear regression according to the number of discussing agents.}
\label{linear_regression_additional}
\end{table}

\subsection{Number of overall agents}
The original experiments were conducted with the number of overall agents $M=100$, but the results could be dependent on the group size.
Therefore, additional experiments were conducted with M=10, 25, and 50 to analyze the results in smaller communities. 
The number of discussing agents was fixed at 5. 
As a result, no particular changes occurred except when $M=10$. 
In the case of $M=10$, because talking with five agents exceeds the majority, it is inevitable that different opinions will be encountered, regardless of the value of $\alpha$. 
As a result, unification occurred in all settings.

\subsection{Initial distribution}
In the original experiments, the distribution of stances was initialized with a uniform distribution of 20\% for each stance but changing the initial distribution could affect the final distribution. 
We conducted additional experiments to investigate this using an initial distribution that assigned ``better to give'' to 60\% of the agents and assigned each of the other stances to 10\% of the agents.
As a result, when $\alpha=0.5$, the stance of agents was unified into ``absolutely must give'' which is the opposite stance from the original experiments. 
When $\alpha=1.0$, it polarized into ``absolutely must give'' and ``absolutely must not give''. 
Although this polarization also happened in the original experiments, ``absolutely must give'' accounted for nearly 80\% in this experiment, showing the opposite trend from the original experiments. 
From this, we can infer that changing the initial distribution can change the final distribution. 
This tendency indicates a security concern that the overall opinion of the AI group could be changed by introducing a large number of AI bots.

\subsection{Presence of reasons}
In the original experiments, the opinion consisted of two elements: stance and reason. 
To investigate how the presence of reasons affects the results, we conducted additional experiments using only stances and excluding the reasons from the inputs and outputs. 
As a result, at $\alpha=0.5$, polarization occurred without the reasons, whereas unification occurred in the original experiments. 
However, the variation in the results was larger than when there were reasons, with two out of three trials resulting in polarization and one trial resulting in unification towards ``better not to give''. 
From this, we can infer that the presence of reasons contributes to the ``stable unification of opinions''.

\subsection{Personas}
ChatGPT can be used to create distinct personalities by embedding a persona into the prompt \cite{pan2023llms}. 
We investigated whether giving each agent a persona would cause changes in the final results. 
We tested two settings in which all agents were given the same persona, ``You are easily swayed by your surroundings and immediately assume that other people's opinions are correct.'' or ``You are a stubborn person and always think you are right.'' 

The final distribution with the ``easily swayed'' personas did not significantly differ from the original results. However, with the ``stubborn'' persona, the final distributions remained almost identical to the initial distribution after 10 turns. Furthermore, the results of the linear regression in Table \ref{linear_regression_persona} show that assigning personas has a significant impact.
In the case of the ``stubborn'' personas, a tendency to stick to one's own stance was observed. In contrast, the ``easily swayed'' personas tended to be influenced by the stances of others. 
From this, we can infer that each agent acts according to its persona, influencing the behavior of the whole group. 

\begin{table}[htb]
\small
\begin{tabular}{|l|l|l|l|l|}
\hline
            & $w_{\mathrm{before}}$ & $\mathrm{w_{around}}$ & $\frac{w_\mathrm{before}}{\mathrm{w_{around}}}$ & coefficient \\ \hline
GPT-4 (stubborn) & 0.999 & 0.00864   & 116   & 0.999       \\ \hline
GPT-4 (neutral)   & 0.724       & 0.526   & 1.38  & 0.957       \\ \hline
GPT-4 (swayed) & 0.203    & 0.895   & 0.227   & 0.940       \\ \hline
\end{tabular}
\caption{The result of linear regression according to personas.}
\label{linear_regression_persona}
\end{table}

\subsection{Order of opinions}
The study on input contexts suggests that language models emphasize the beginning and end of the prompt \cite{liu2023lost}. 
Similarly, where the opinion of each discussing agent is described in the prompt might influence the agent's stance after the discussion. 
Based on this hypothesis, we measured the correlation between the order of the discussing agents and the stance after the discussion. 
However, no significant relationship was observed between the order of agents and the results. 
Therefore, the order of the opinions did not significantly impact the results.

\subsection{Frequency penalty}
ChatGPT has a parameter called the frequency penalty, which imposes a penalty on token reuse. 
In the original experiments, we used the default value of 0, but we conducted additional experiments by changing this value to 1.0 and -1.0. 
However, no particular influence was observed in the final results.

\subsection{Summary}
This research involved numerous parameters. 
Through additional experiments, we deduced that the number of discussing agents, initial distribution, presence of reasons, and persona have a significant impact. 
By contrast, the group size, order of opinions, and frequency penalty do not influence the results. 
Parameters with a significant impact indicate vulnerabilities when viewed from the attacker's perspective and thus should be monitored in terms of security. 
In addition, with respect to topic and language, experiments on more multilingual and broader topics would provide more detail on the factors that influence polarization.

\section{Discussion and Conclusion}
In this study, we verified whether a group of autonomous AI agents based on generative AI could cause polarization under the condition of an echo chamber. 
We proposed a new framework for simulating the polarization of AI agents, and the results of the simulation demonstrated that agents based on ChatGPT can polarize when in an echo chamber.
The analysis of the opinion transitions revealed that this polarization can be attributed to the strong ability of ChatGPT to understand prompts and update its own opinion by considering both its own and the surrounding opinions.
Moreover, through additional experiments, we identified factors that strongly influence polarization, such as the persona.

We note that this study does not indicate what distribution of opinions is desirable for AI agents. 
A diversity of opinions on some topics is desirable.
However, for other topics such as ``It is good to discriminate against minorities,'' it would not benefit society to have an even split between agreement, neutral, and disagreement. 
The ideal opinion distributions among AI agents depend on each topic and culture. 
They must be decided by discussions within each society. 

A limitation of this study is that we modeled each agent and its interactions in a simplified manner. 
In reality, one's opinions are formed not in organized discussions but through daily exposure to news and casual conversations with others. 
Future research will include simulations based on a detailed modeling of how AI agents will be used in reality. 

\clearpage

\bibliographystyle{ACM-Reference-Format}
\bibliography{main}
\end{document}